\def\year{2018}\relax

\documentclass[letterpaper]{article}
\usepackage{aaai18}
\usepackage{times}
\usepackage{lipsum}
\usepackage{helvet}
\usepackage{courier}
\usepackage{mathtools}
\usepackage{amsmath}
\usepackage{latexsym}
\usepackage{float}
\usepackage{graphicx}
\usepackage{algorithm}
\usepackage{algpseudocode}
\usepackage{amssymb}
\usepackage{array}
\usepackage{multirow}
\usepackage{booktabs}
\usepackage{siunitx}
\usepackage{url}
\usepackage{tablefootnote}
\usepackage{amsthm}
\usepackage{amssymb}
\usepackage{pifont}
\usepackage{array}
\usepackage{xcolor}
\usepackage{soul}
\usepackage{booktabs} 
\usepackage{lipsum}
\usepackage{multirow}
\usepackage{array}
\usepackage{courier}
\usepackage{mathtools}
\usepackage{subcaption}
\usepackage{amsfonts}
\usepackage{tikz}
\usepackage{tabularx}
\usepackage{mathtools}
\usepackage{xcolor}
\usepackage{coloremoji}

\definecolor{8red}{HTML}{A70000}
\definecolor{6red}{HTML}{FF0000}
\definecolor{4red}{HTML}{FF5252}
\definecolor{2red}{HTML}{FF7B7B}
\definecolor{1red}{HTML}{FFBABA}

\DeclareRobustCommand{\hltwo}[1]{{\sethlcolor{2red}\hl{#1}}}

\title{\textsc{CoupleNet:} Paying Attention to Couples with Coupled Attention for Relationship Recommendation}

\author{Yi Tay\textsuperscript{$1$}, Luu Anh Tuan\textsuperscript{$2$} \and Siu Cheung Hui\textsuperscript{$3$}\\
\textsuperscript{$1,3$}\:Nanyang Technological University \\ School of Computer Science and Engineering, Singapore \\
\textsuperscript{$2$}\:Institute for Infocomm Research, Singapore \\}
\date{}

\begin{document}
\maketitle
\begin{abstract}
  Dating and romantic relationships not only play a huge role in our personal lives but also collectively influence and shape society. Today, many romantic partnerships originate from the Internet, signifying the importance of technology and the web in modern dating. In this paper, we present a text-based computational approach for estimating the relationship compatibility of two users on social media.  Unlike many previous works that propose reciprocal recommender systems for online dating websites, we devise a distant supervision heuristic to obtain real world couples from social platforms such as Twitter. Our approach, the \textsc{CoupleNet} is an end-to-end deep learning based
estimator that analyzes the social profiles of two users and subsequently performs a similarity match between the users. Intuitively, our approach performs both user profiling and match-making within a unified end-to-end framework. \textsc{CoupleNet} utilizes hierarchical recurrent neural models for learning representations of user profiles and subsequently coupled attention mechanisms to fuse information aggregated from two users.
To the best of our knowledge, our approach is the first data-driven deep learning approach for our novel relationship recommendation problem.
 We benchmark
our \textsc{CoupleNet} against several machine learning and deep learning baselines. Experimental results show that our approach outperforms
all approaches significantly in terms of precision. Qualitative analysis shows that our model is capable of also producing explainable results to users.
\end{abstract}

\section{Introduction}
The social web has become a common means for seeking romantic companionship, made evident by the wide assortment of online dating sites that are available on the Internet. As such, the notion of relationship recommendation systems is not only interesting but also highly applicable. This paper investigates the possibility and effectiveness of a deep learning based relationship recommendation system. An overarching research question is whether modern artificial intelligence (AI) techniques, given social profiles, can successfully approximate successful relationships and measure the relationship compatibility of two users.

Prior works in this area \cite{Xia:2015:RRS:2808797.2809282,ICWSM148061,IAAI148187,Xia:2015:RRS:2808797.2809282} have been mainly considered the `online dating recommendation' problem, i.e., focusing on the reciprocal domain of dating social networks (DSN) such as Tinder and OKCupid. While the functionality and mechanics of dating sites differ across the spectrum, the main objective is usually to facilitate communication between users, who are explicitly seeking relationships. Another key characteristic of many DSNs is the functionality that enables a user to express interest to another user, e.g., swiping right on Tinder. Therefore, many of prior work in this area focus on reciprocal recommendation, i.e., predicting if two users will \textit{like} or \textit{text} each other. Intuitively, we note that likes and replies on DSNs are not any concrete statements of compatibility nor evidence of any long-term relationship. For instance, a user may have many reciprocal matches on Tinder but eventually form meaningful friendships or relationships with only a small fraction.

Our work, however, focuses on a seemingly similar but vastly different problem. Instead of relying on reciprocal signals from DSNs, our work proposes a novel distant supervision scheme, constructing a dataset of real world couples from regular\footnote{We define regular social networks (RSN) as any social network that is not primarily a DSN, e.g., Facebook, Twitter.} social networks (RSN). Our distant supervision scheme is based on Twitter, searching for tweets such as \textit{`good night baby love you 😘'} and \textit{`darling i love you so much 💓'} to indicate that two users are in a stable and loving relationship (at least at that time). Using this labeled dataset, we train a distant supervision based learning to rank model to predict relationship compatibility between two users using their social profiles. The key idea is that social profiles contain cues pertaining to personality and interests that may be a predictor if whether two people are romantically compatible. Moreover, unlike many prior works that operate on propriety datasets \cite{ICWSM148061,IAAI148187,Xia:2015:RRS:2808797.2809282}, our dataset is publicly and legally obtainable via the official Twitter API. In this work, we construct the first public dataset of approximately 2 million tweets for the task of relationship recommendation.

Another key advantage is that our method trains on regular social networks, which spares itself from the inherent problems faced by DSNs, e.g., deceptive self-presentation, harassment, bots, etc. \cite{Masden:2015:URC:2702123.2702417}. More specifically, self-presented information on DSNs might be inaccurate with the sole motivation of appearing more attractive \cite{Toma:2010:RLL:1718918.1718921,hancock2007truth}. In our work, we argue that measuring the compatibility of two users on RSN might be more suitable, eliminating any potential explicit self-presentation bias. Intuitively, social posts such as tweets can reveal information regarding personality, interests and attributes \cite{ICWSM1715681,Wei:2017:BWP:3018661.3018717}.

Finally, we propose \textsc{CoupleNet}, an end-to-end deep learning based architecture for estimating the compatibility of two users on RSNs. \textsc{CoupleNet} takes the social profiles of two users as an input and computes a compatibility score. This score can then be used to serve a ranked list to users and subsequently embedded in some kind of `who to follow' service. \textsc{CoupleNet} is characterized by its Coupled Attention, which learns to pay attention to parts of a user's profile dynamically based on the current candidate user. \textsc{CoupleNet} also does not require any feature engineering and is a proof-of-concept of a completely text-based relationship recommender system. Additionally, \textsc{CoupleNet} is also capable of providing explainable recommendations which we further elaborate in our qualitative experiments.

\subsection{Our Contributions}
This section provides an overview of the main contributions of this work.
\begin{itemize}
\item We propose a novel problem of \textit{relationship recommendation} (RSR). Different from the reciprocal recommendation problem on DSNs, our RSR task operates on regular social networks (RSN), estimating long-term and serious relationship compatibility based on social posts such as tweets.
\item We propose a novel distant supervision scheme to construct the first publicly available (distributable in the form of tweet ids) dataset for the RSR task. Our dataset, which we call the \textsc{LoveBirds2M} dataset consists of approximately 2 million tweets.
\item We propose a novel deep learning model for the task of RSR. Our model, the \textsc{CoupleNet} uses hierarchical Gated Recurrent Units (GRUs) and coupled attention layers to model the interactions between two users. To the best of our knowledge, this is the first deep learning model for both RSR and reciprocal recommendation problems.
\item We evaluate several strong machine learning and neural baselines on the RSR task. This includes the recently proposed DeepCoNN (\textit{Deep Co-operative Neural Networks}) \cite{zheng2017joint} for item recommendation. \textsc{CoupleNet} significantly outperforms DeepCoNN with a $200\%$ relative improvement in precision metrics such as Hit Ratio (HR@N). Overall findings show that a text-only deep learning system for RSR task is plausible and reasonably effective.
\item We show that \textsc{CoupleNet} produces explainable recommendation by analyzing the attention maps of the coupled attention layers.
\end{itemize}

\section{Related Work}
In this section, we review existing literature that is related to our work.
\subsection{Reciprocal and Dating Recommendation}

Prior works on online dating recommendation \cite{Xia:2015:RRS:2808797.2809282,Tu:2014:ODR:2567948.2579240,IAAI148187,DBLP:conf/ijcai/AkehurstKYPKR11} mainly focus on designing systems for dating social networks (DSN), i.e., websites whereby users are on for the specific purpose of finding a potential partner. Moreover, all existing works have primarily focused on the notion of reciprocal relationships, e.g., a successful signal implied a two way signal (likes or replies) between two users.

Tu et al. \cite{Tu:2014:ODR:2567948.2579240} proposed a recommendation system based on Latent Dirichlet Allocation (LDA) to match users based on messaging and conversational history between users. Xia et al. \cite{Xia:2015:RRS:2808797.2809282,ICWSM148061} cast the dating recommendation problem into a link prediction task, proposing a graph-based approach based on user interactions. The CCR (Content-Collaborative Reciprocal Recommender System) \cite{DBLP:conf/ijcai/AkehurstKYPKR11} was proposed by Akehurtst et al. for the task of reciprocal recommendation, utilizing content-based features (user profile similarity) and collaborative filtering features (user-user interactions). However, all of their approaches operate on a propriety dataset obtained via collaboration with online dating sites. This hinders research efforts in this domain.

 Our work proposes a different direction from the standard reciprocal recommendation (RR) models. The objective of our work is fundamentally different, i.e., instead of finding users that might reciprocate to each other, we learn to functionally approximate the essence of a good (possibly stable and serious) relationship, learning a compatibility score for two users given their regular social profiles (e.g., Twitter). To the best of our knowledge, our work is the first to build a relationship recommendation model based on a distant supervision signal on real world relationships. Hence, we distinguish our work from all existing works on online dating recommendation.

Moreover, our dataset is obtained legally via the official twitter API and can be distributed for future research. Unlike prior work \cite{Xia:2015:RRS:2808797.2809282} which might invoke privacy concerns especially with the usage of conversation history, the users employed in our study have public twitter feeds. We note that publicly available twitter datasets have been the cornerstone of many scientific studies especially in the fields of social science and natural language processing (NLP).

Across scientific literature, several other aspects of online dating have been extensively studied. Nagarajan and Hearst \cite{DBLP:conf/icwsm/NagarajanH09} studied self-presentation on online dating sites by specifically examining language on dating profiles. Hancock et al. presented an analysis on deception and lying on online dating profiles \cite{hancock2007truth}, reporting that at least $50\%$ of participants provide deceptive information pertaining to physical attributes such as height, weight or age. Toma et al. \cite{Toma:2010:RLL:1718918.1718921} investigated the correlation between linguistic cues and deception on online dating profiles. Maldeniya et al. \cite{ICWSM1715634} studied how textual similarity between user profiles impacts the likelihood of reciprocal behavior. A recent work by Cobb and Kohno \cite{DBLP:conf/www/CobbK17} provided an extensive study which tries to understand users’ privacy preferences and practices
in online dating.

Finally, \cite{garimella2014love} studied the impacts of relationship breakups on Twitter, revealing many crucial insights pertaining to the social and linguistic behaviour of couples that have just broken up. In order to do so, they collect likely couple pairs and monitor them over a period of time. Notably, our data collection procedure is reminscent of theirs, i.e., using keyword-based filters to find highly likely couple pairs. However, their work utilizes a second stage crowdworker based evaluation to check for breakups.

\subsection{User Profiling and Friend Recommendation}

Our work is a cross between user profiling and user match-making systems. An earlier work, \cite{DBLP:conf/sigir/DiazMA10} proposed a gradient-boosted learning-to-rank model for match-making users on a dating forum. While the authors ran experiments on a dating service website, the authors drew parallels with other match-making services such as job-seeking forums. The user profiling aspect in our work comes from the fact that we use social networks to learn user representations. As such, our approach performs both user profiling and then match-making within an end-to-end framework. \cite{Wei:2017:BWP:3018661.3018717} proposed a deep learning personality detection system which is trained on social posts on Weibo and Twitter. \cite{ICWSM1715681} proposed a Twitter personality detection system based on machine learning models. \cite{DBLP:conf/acl/BentonAD16} learned multi-view embeddings of Twitter users using canonical correlation analysis for friend recommendation. From an application perspective, our work is also highly related to `People you might know' or `who to follow' (WTF) services on RSNs \cite{Gupta:2013:WFS:2488388.2488433} albeit taking a romantic twist.  In practical applications, our RSN based relationship recommender can either be deployed as part of a WTF service, or to increase the visibility of the content of users with high compatibility score.

\subsection{Deep Learning and Collaborative Ranking}
One-class collaborative filtering (also known as collaborative ranking) \cite{hu2008collaborative} is a central research problem in IR. In general, deep learning \cite{He:2017:NCF:3038912.3052569,Tay:2018:LRM:3178876.3186154,zhang2018neurec} has also been recently very popular for collaborative ranking problems today. However, to the best of our knowledge, our work is the first deep learning based approach for the online dating domain. \cite{zhang2017deep} provides a comprehensive overview of deep learning methods for CF. Notably, our approach also follows the neural IR approach which is mainly concerned with modeling document-query pairs \cite{DBLP:conf/sigir/SeverynM15,DBLP:conf/sigir/TayPLH17,tay2017cross} or user-item pairs \cite{zheng2017joint,DBLP:journals/corr/abs-1801-09251} since we deal with the textual domain. Finally, our work leverages recent advances in deep learning, namely Gated Recurrent Units \cite{DBLP:journals/corr/ChoMGBSB14} and Neural Attention \cite{yang2016hierarchical,luong2015effective,bahdanau2014neural}. The key idea of neural attention is to learn to attend to various segments of a document, eliminating noise and emphasizing the important segments for prediction.

\newtheorem{Definition}{Definition}[section]{}

\section{Problem Definition and Notation}

In this section, we introduce the formal problem definition of this work.
\begin{Definition}
Let $U$ be the set of Users. Let $s_i$ be the social profile of user $i$ which is denoted by $u_i \in U$. Each social profile $s_i \in S$ contains $\eta$ documents. Each document $d_i \in s_i$ contains a maximum of $L$ words. Given a user $u_i$ and his or her social profile $s_i$, the task of the Relationship Recommendation problem is to produce a ranked list of candidates based on a computed relevance score $F(s_i, s_j)$ where $s_j$ is the social profile of the candidate user $u_j$. $F(.)$ is a parameterized function.
\end{Definition}

There are mainly three types of learning to rank methods, namely pointwise, pairwise and list-wise. Pointwise considers each user pair individually, computing a relevance score solely based on the current sample, i.e., binary classification. Pairwise trains via noise constrastive estimation, which often minimizes a loss function like the margin based hinge loss. List-wise considers an entire list of candidates and is seldom employed due to the cumbersome constraints that stem from implementation efforts. Our proposed \textsc{CoupleNet} employs a pairwise paradigm. The intuition for this is that, relationship recommendation is considered very sparse and has very imbalanced classes (for each user, only one ground truth exists). Hence, training binary classification models suffers from class imbalance. Moreover, the good performance of pairwise learning to rank is also motivated by our early experiments.

\section{The Love Birds Dataset}
\begin{figure}[ht]
  \includegraphics[width=0.38\textwidth]{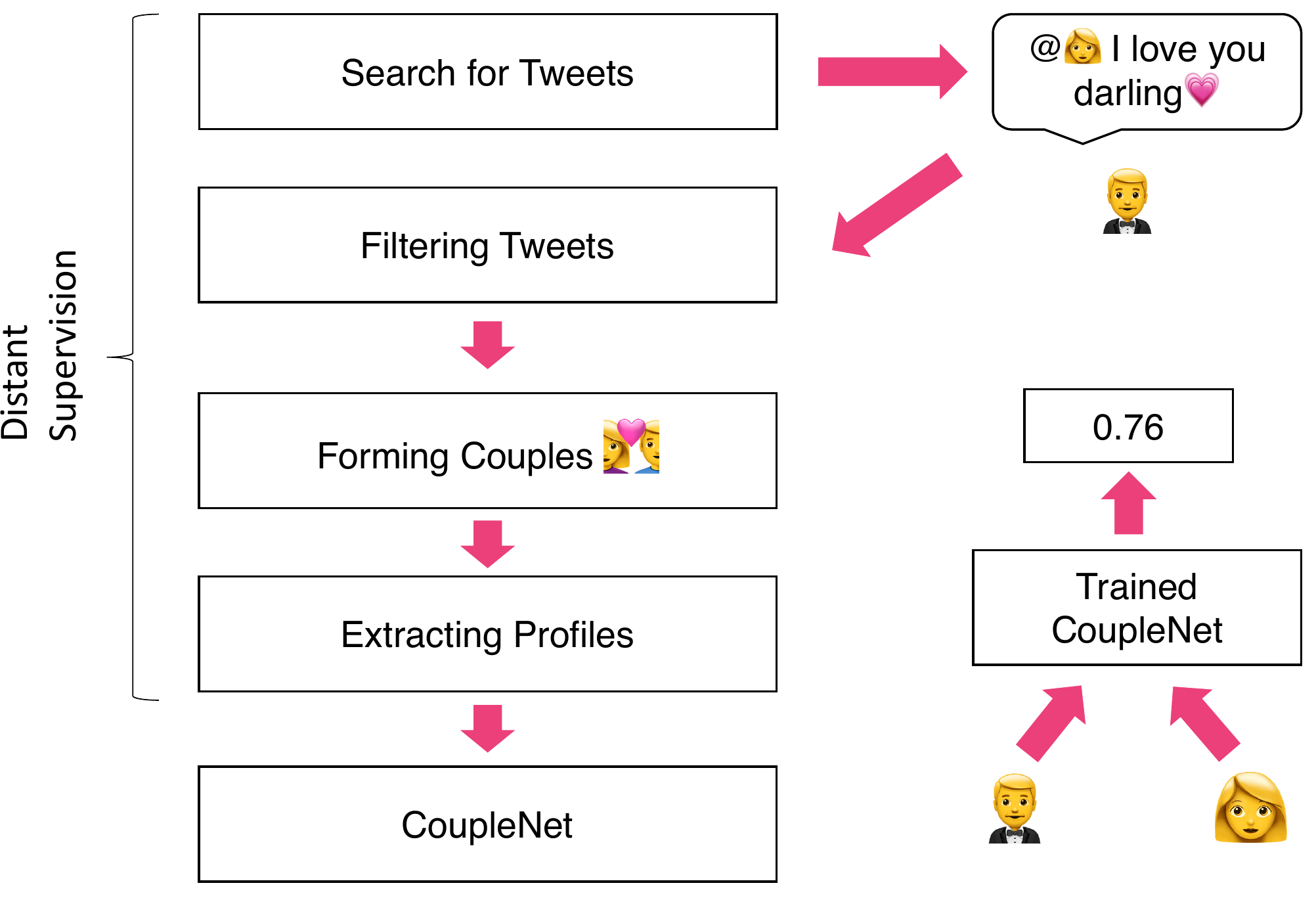}
  \caption{Overview of our distant supervision and deep learning approach for relationship recommendation.}\label{overview}
  \end{figure}

Since there are no publicly available datasets for training relationship recommendation models, we construct our own. The goal is to construct a list of user pairs in which both users are in relationship. Our dataset is
constructed via distant supervision from Twitter. We call this dataset the \textit{Love Birds} dataset. This not only references the metaphorical meaning of the phrase `love birds' but also deliberately references the fact that
the Twitter icon is a bird. This section describes the construction of our dataset\footnote{To facilitate further research, our dataset will be released at \url{https://github.com/vanzytay/ICWSM18_LB2M}. Distribution will come in the form of tweet IDs and labels, to adhere to the regulations of the Twitter public API. }. Figure \ref{overview} describes the overall process of our distant supervision framework.

\subsection{Distant Supervision}
Using the Twitter public API, we collected tweets with emojis contains the keyword \textit{`heart'} in its description. The key is to find tweets where a user expresses love to another user. We observed that there are countless tweets such as \textit{`good night baby love you 😘'} and \textit{`darling i love you so much 💓'} on Twitter. As such, the initial list of tweets is crawled by watching heart and love-related emojis, e.g.,  😘, 💓, 💖 etc. By collecting tweets containing these emojis, we form our initial candidate list of couple tweets (tweets in which two people in a relationship send to each other). Through this process, we collected 10 million tweets over a span of a couple of days. Each tweet will contain a sender and a target (the user mentioned and also the target of affection).
\subsubsection{Keyword Filtering}
We also noticed that the love related emojis do not necessarily imply a romantic relationship between two users. For instance, we noticed that a large percentage of such tweets are affection towards family members. Given the large corpus of candidates, we can apply a stricter filtering rule to obtain true couples. To this end, we use a ban list of words such as 'bro', 'sis', `dad', `mum' and apply regular expression based filtering on the candidates. We also observed a huge amount of music related tweets, e.g., `I love this song so much 💓!'. Hence, we also included music-related keywords such as `perform', `music', `official' and `song'. Finally, we also noticed that people use the heart emoji frequently when asking for someone to follow them back. As such, we also ban the word `follow'.
\subsubsection{User-based Filtering}
We further restricted tweets to contain only a single mention. Intuitively, mentioning more than one person implies a group message rather than a couple tweet. We also checked if one user has a much higher follower count over the other user. In this case, we found that this is because people send love messages to popular pop idols (we found that  a huge bulk of crawled tweets came from fangirls sending love message to @harrystylesofficial). Any tweet with a user containing more than 5K followers is being removed from the candidate list.
\subsection{Forming Couple Pairs}
Finally, we arrive at 12K tweets after aggressive filtering. Using the 12K `cleaned' couple tweets, we formed a list of couples.  We sorted couples in alphabetical order, i.e., (clara, ben) becomes (ben, clara) and removed duplicate couples to ensure that there are no `bidirectional' pairs in the dataset. For each user on this list, we crawled their timeline and collected 200 latest tweets from their timeline. Subsequently, we applied further preprocessing to remove explicit couple information. Notably, we do not differentiate between male and female users (since twitter API does not provide this information either). The signal for distant supervision can be thought of as an explicit signal which is commonplace in recommendation problems that are based on explicit feedback (user ratings, reviews, etc.). In this case, an act (tweet) of love / affection is the signal used. We call this explicit couple information.
\subsubsection{Removing Additional Explicit Couple Information}
To ensure that there are no \textit{additional} explicit couple information in each user's timeline, we removed all tweets with any words of affection (heart-related emojis, `love', `dear', etc.). We also masked all mentions with the @USER symbol. This is to ensure that there is no explicit leak of signals in the final dataset. Naturally, a more accurate method is to determine the date in which users got to know each other and then subsequently construct timelines based on tweets prior to that date. Unfortunately, there is no automatic and trivial way to easily determine this information. Consequently, a fraction of their timeline would possibly have been tweeted when the users have already been together in a relationship. As such, in order to remove as much 'couple' signals, we try our best to mask such information.
\subsection{Why Twitter?}
Finally, we answer the question of why Twitter was chosen as our primary data source. One key desiderata was that the data should be public, differentiating ourselves from other works that use proprietary datasets \cite{Xia:2015:RRS:2808797.2809282,Tu:2014:ODR:2567948.2579240}. In designing our experiments, we considered two other popular social platforms, i.e., Facebook and Instagram. Firstly, while Facebook provides explicit relationship information, we found that there is a lack of personal, personality-revealing posts on Facebook. For a large majority of users, the only signals on Facebook mainly consist of shares and likes of articles. The amount of original content created per user is extremely low compared to Twitter whereby it is trivial to obtain more than 200 tweets per user. Pertaining to Instagram, we found that posts are also generally much sparser especially in regards to frequency, making it difficult to amass large amounts of data per user. Moreover, Instagram adds a layer of difficulty as Instagram is primarily multi-modal. In our Twitter dataset, we can easily mask explicit couple information by keyword filters. However, it is non-trivial to mask a user's face on an image. Nevertheless, we would like to consider Instagram as an interesting line of future work.
\subsection{Dataset Statistics}
Our final dataset consists of 1.858M tweets (200 tweets per user). The total number of users is 9290 and 4645 couple pairs. The couple pairs are split into training, testing and development with a 80/10/10 split. The total vocabulary size (after lowercasing) is 2.33M. Ideally, more user pairs could be included in the dataset. However, we also note that the dataset is quite large (almost 2 million tweets) already, posing a challenge for standard hardware with mid-range graphic cards. Since this is the first dataset created for this novel problem, we leave the construction of a larger benchmark for future work.

\section{Our Proposed Approach}
\begin{figure}[ht]
  \includegraphics[width=0.46\textwidth]{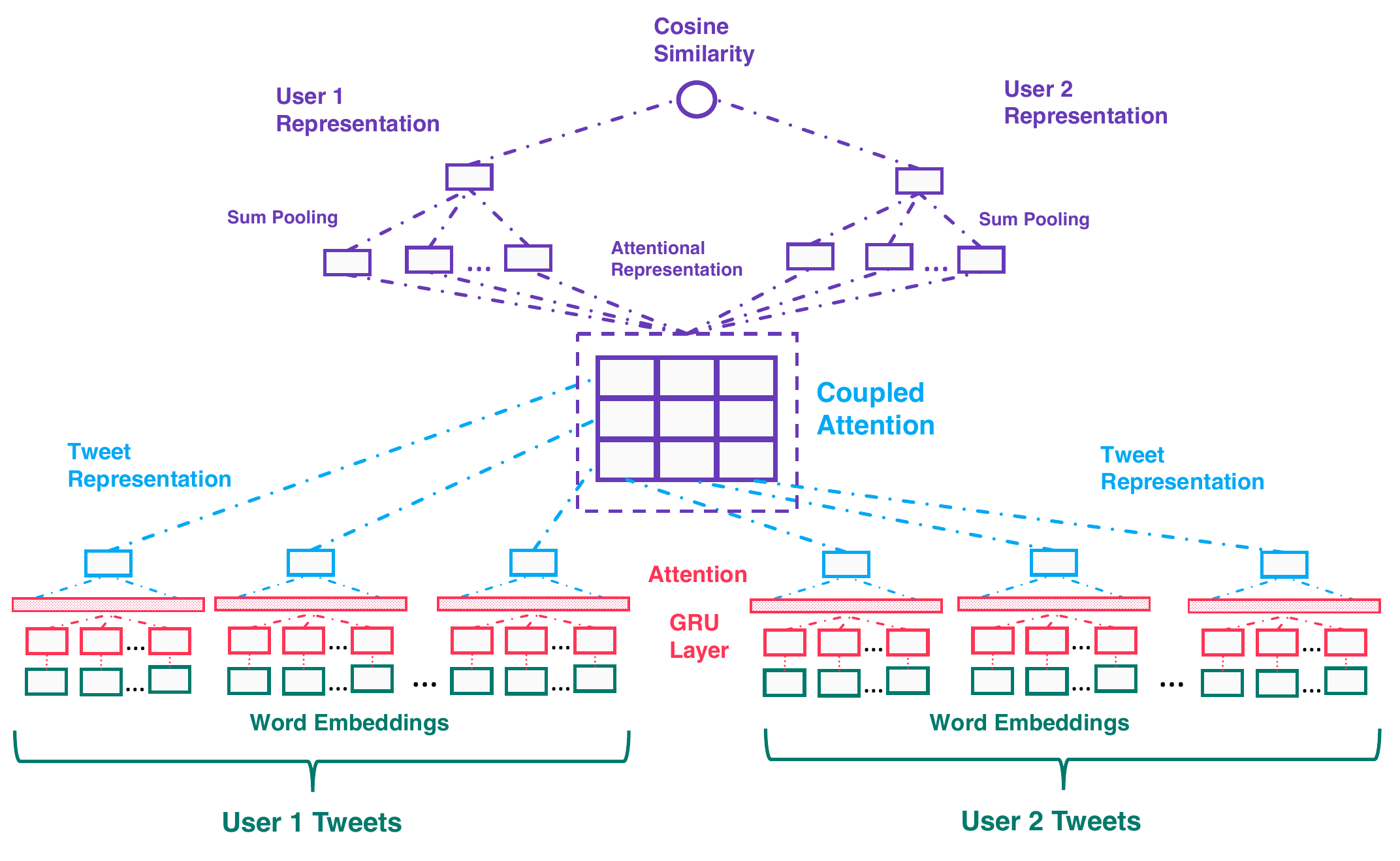}
  \caption{Overview of \textsc{CoupleNet} model architecture illustrating the computation of similarity score for User 1 and User 2. Negative sampling side of the network is omitted due to lack of space. }\label{modelarch}
  \end{figure}

In this section, we introduce our deep learning architecture - the \textsc{CoupleNet}. Overall, our neural architecture is a hierarchical recurrent model \cite{yang2016hierarchical}, utilizing multi-layered attentions at different hierarchical levels. An overview of the model architecture is illustrated in Figure \ref{modelarch}. There are two sides of the network, one for each user. Our network follows a `Siamese' architecture, with shared parameters for each side of the network. A single data input to our model comprises user pairs ($U1, U2$) (couples) and ($U1, U3$) (negative samples). Each user has $K$ tweets each with a maximum length of $L$. The value of $K$ and $L$ are tunnable hyperparameters.

\subsection{Embedding Layer}
For each user, the inputs to our network are a matrix of indices, each corresponding to a specific word in the dictionary. The embedding matrix $\textbf{W} \in \mathbb{R}^{d \times |V|}$ acts as a look-up whereby each index selects a $d$ dimensional vector, i.e., the word representation. Thus, for each user, we have $K \times L$ vectors of dimension size $d$. The embedding layer is shared for all users and is initialized with pretrained word vectors.

\subsection{Learning Tweet Representations}
For each user, the output of the embedding layer is a tensor of shape $K \times L \times d$. We pass each tweet through a recurrent neural network. More specifically, we use Gated Recurrent Units (GRU) encoders with attentional pooling to learn a $n$ dimensional vector for each tweet.
\subsubsection{Gated Recurrent Units (GRU)}

The GRU accepts a sequence of vectors and recursively composes each input vector into a hidden state. The recursive operation of the GRU is defined as follows:
\begin{align*}
z_t &= \sigma (W_z x_t + U_z h_{t-1} + b_z) \\
r_t &= \sigma (W_r x_t + U_r h_{t-1} + b_r) \\
\hat{h_t} &= tanh (W_h \: x_t + U_h (r_t  h_{t-1}) + b_h) \\
h_t &= z_t \: h_{t-1} + (1-z_t) \: \hat{h_t}
\end{align*}
where $h_t$ is the hidden state at time step $t$, $z_t$ and $r_t$ are the update gate and reset gate at time step $t$ respectively. $\sigma$ is the sigmoid function. $x_t$ is the input to the GRU unit at time step $t$. Note that time step is analogous to parsing a sequence of words sequentially in this context. $W_z, W_r \in \mathbb{R}^{d \times n}, W_h \in \mathbb{R}^{n \times n}$ are parameters of the GRU layer.
\subsubsection{Tweet-level Attention}
The output of each GRU is a sequence of hidden vectors $h_1, h_2 \cdots h_L \in \textbf{H}$, where $\textbf{H} \in \mathbb{R}^{L \times n}$. Each hidden vector is $n$ dimensions, which corresponds to the parameter size of the GRU. To learn a single $n$ dimensional vector, the last hidden vector $h_L$ is typically considered. However, a variety of pooling functions such as the average pooling, max pooling or attentional pooling can be adopted to learn more informative representations. More specifically, neural attention mechanisms are applied across the matrix $\textbf{H}$, learning a weighted representation of all hidden vectors. Intuitively, this learns to select more informative words to be passed to subsequent layers, potentially reducing noise and improving model performance.
\begin{align*}
\textbf{Y} = \text{tanh}(W_y \: \textbf{H}) \:\:;\:\: a= \text{softmax}(w^{\top} \: \textbf{Y}) \:\:;\:\: r = \textbf{H}\: a^{\top}
\end{align*}
where $W_y \in \mathbb{R}^{n \times n}, w \in \mathbb{R}^{n}$ are the parameters of the attention pooling layer. The output $r \in \mathbb{R}^{n}$ is the final vector representation of the tweet. Note that the parameters of the attentional pooling layer are shared across all tweets and across both users.

\subsection{Learning User Representations}
Recall that each user is represented by $K$ tweets and for each tweet we have a $n$ dimensional vector. Let $t^i_1, t^i_2 \cdots t^i_K$ be all the tweets for a given user $i$. In order to learn a fixed $n$ dimensional vector for each user, we require a pooling function across each user's tweet embeddings. In order to do so, we use a Coupled Attention Layer that learns
to attend to U1 based on U2 (and vice versa). Similarly, for the negative sample, coupled attention is applied to (U1, U3) instead. However, we only describe the operation of (U1, U2) for the sake of brevity.

\subsubsection{Coupled Attention} The key intuition behind the coupled attention layer is to learn attentional representations of U1 with respect to U2 (and vice versa). Intuitively, this compares each tweet of U1 with each tweet of U2 and learns to weight each tweet based on this grid-wise comparison scheme. Let U1 and U2 be represented by a sequence of $K$ tweets (each of which is a $n$ dimensional vector) and let $T_1, T_2 \in \mathbb{R}^{k \times n}$ be the tweet matrix for U1 and U2 respectively. For each tweet pair ($t^{1}_i, t^{2}_j$), we utilize a feed-forward neural network to learn a similarity score between each tweet. As such, each value of the similarity grid is computed:
\begin{equation}
s_{ij} = W_{c} \: [t^{1}_i; t^{2}_j] + b_c
\end{equation}
where $W_c \in \mathbb{R}^{n \times 1}$ and $b_c \in \mathbb{R}^{1}$ are parameters of the feed-forward neural network. Note that these parameters are shared across all tweet pair comparisons. The score $s_{ij}$ is a scalar value indicating the similarity between tweet $i$ of U1 and tweet $j$ of U2.
\subsubsection{Aggregating Strong Signals}
Given the similarity matrix $\textbf{S} \in \mathbb{R}^{K \times K}$, the strongest signals across each dimension are aggregated using max pooling. For example, by taking a max over the columns of \textbf{S}, we regard the importance of tweet $i$ of U1 as the strongest influence it has over all tweets of U2. The result of this aggregation is two $K$ length vectors which are used to attend over the original sequence of tweets. The following operations describe the aggregation functions:
\begin{align}
a^{row} = \text{smax}(\max_{row} \textbf{S}) \:\:\:\text{and}\:\:\: a^{col} = \text{smax}(\max_{col} \textbf{S})
\end{align}
where $a^{row}, a^{col} \in \mathbb{R}^{K}$ and smax is the softmax function. Subsequently, both of these vectors are used to attentively pool the tweet vectors of each user.
\begin{align*}
u_1 = T_1 \: a^{col} \:\:\text{and}\:\:u_2 = T_2 \: a^{row}
\end{align*}
where $u_1, u_2 \in \mathbb{R}^{n}$ are the final user representations for U1 and U2.
\subsection{Learning to Rank and Training Procedure}
Given embeddings $u_1, u_2, u_3$, we introduce our similarity modeling layer and learning to rank objective. Given $u_1$ and $u_2$, the similarity between each user pair is modeled as follows:
\begin{equation}
s(u_1, u_2) = \frac{u_i \cdot u_2}{|u_1| |u_2|}
\end{equation}
which is the cosine similarity function. Subsequently, the pairwise ranking loss is optimized. We use the margin-based hinge loss to optimize our model.
\begin{equation}
J =  \max \{0, \lambda - s(u_1,u_2) + s(u_1, u_3) \}
\end{equation}
where $\lambda$ is the margin hyperparameter, $s(u_1, u_2)$ is the similarity score for the ground truth (true couples) and $s(u_1, u_3)$ is the similarity score for the negative sample. This function aims to discriminate between couples and non-couples by increasing the margin between the ranking scores of these user pairs. Parameters of the network can be optimized efficiently with stochastic gradient descent (SGD).

\section{Empirical Evaluation}
Our experiments are designed to answer the following Research Questions (\textbf{RQ}s).
\begin{itemize}
\item \textbf{RQ1} - How well are machine learning and deep learning methods able to learn, predict, recommend relationships just based on linguistic information from social profiles? Are the romantic compatibility of two people predictable just based on textual information?
\item \textbf{RQ2} - Does the amount of information (number of tweets per user) affect the ability to recommend relationships?
\item \textbf{RQ3} - Are we able to derive any insight on how these models are learning to recommend relationships? Are attention models able to produce explainable relationship recommendations?
\end{itemize}

\subsection{Experimental Setup}
All empirical evaluation is conducted on our LoveBirds dataset which has been described earlier. This section describes the evaluation metrics used and evaluation procedure.
\subsubsection{Evaluation Metrics}
Our problem is posed as a learning-to-rank problem. As such, the evaluation metrics used are as follows:
\begin{itemize}
\item \textbf{Hit Ratio @N} is the ratio of test samples which are correctly retrieved within the top $N$ users. We evaluate on
$N=10,5,3$.
\item \textbf{Accuracy} is the number of test samples that have been correctly ranked in the top position.
\item \textbf{Mean Reciprocal Rank (MRR)} is a commonly used information retrieval metric. The reciprocal rank of a single test sample is the multiplicative inverse of the rank. The MRR is computed by $\frac{1}{Q} \sum^{|Q|}_{i=1} \frac{1}{rank_i}$.
\item \textbf{Mean Rank} is the average rank of all test samples.
\end{itemize}
\subsubsection{Evaluation Procedure}
Our experimental procedure samples $100$ users per test sample and ranks the golden sample amongst the $100$ negative samples.

\subsubsection{Algorithms Compared}
In this section, we discuss the algorithms and baselines compared. Notably, there are no established benchmarks for this new problem. As such, we create 6 baselines to compare against our proposed \textsc{CoupleNet}.
\begin{itemize}
\item \textbf{RankSVM (Tf-idf)} - This model is a RankSVM (Support Vector Machine) trained on tf-idf vectors. This model is known to be a powerful vector space model (VSM) baseline. The feature vector of each user is a $k$ dimensional vector, representing the top-$k$ most common n-grams. The n-gram range is set to (1,3) and $k$ is set to 5000 in our experiments. Following the original implementation, the kernel of RankSVM is a linear kernel.
\item \textbf{RankSVM (Embed)} - This model is a RankSVM model trained on pretrained (static, un-tuned) word embeddings. For each user pair, the feature vector is the sum of all words of both users.
\item \textbf{MLP (Embed)} - This is a Multi-layered Perceptron (MLP) model that learns to non-linearly project static word embedding. Each word embedding is projected using 2 layered MLP with ReLU activations. The user representation is the sum of all transformed word embeddings.
\item \textbf{DeepCoNN (Deep Co-operative Neural Networks)} \cite{zheng2017joint} is a convolutional neural network (CNN). CNNs learn n-gram features by sliding weights across an input. In this model, all of a user's tweets are concatenated and encoded into a $d$ dimensional vector via a convolutional encoder. We use a fixed filter width of $3$. DeepCoNN was originally proposed for item recommendation task using reviews. In our context, we adapt the DeepCoNN for our RSR task (tweets are analogous to reviews). Given the different objectives (MSE vs ranking), we also switch\footnote{In our problem, we found that the FM layer significantly degraded performance.} the factorization machine (FM) layer for the cosine similarity.
The number of filters is $100$. A max pooling layer is used to aggregate features.
\item \textbf{Baseline Gated Recurrent Unit (GRU)} - We compare with a baseline GRU model. Similar to the DeepCoNN model, the baseline GRU considers a user to be a concatenation of all the user's tweets. The size of the recurrent cell is $100$ dimensions.
\item \textbf{Hierarchical GRU (H-GRU)} - This model learns user representations by first encoding each tweet with a GRU encoder. The tweet embedding is the last hidden state of the GRU. Subsequently, all tweet embeddings are summed. This model serves as an ablation baseline of our model, i.e., removing all attentional pooling functions.
\end{itemize}

\subsubsection{Implementation Details}
All models were implemented in Tensorflow on a Linux machine. For all neural network models, we follow a \textit{Siamese} architecture (shared parameters for both users) and mainly vary the neural encoder. The cosine ranking function and hinge loss are then used to optimize all models. We train all models with the Adam \cite{DBLP:journals/corr/KingmaB14} optimizer with a learning rate of $10^{-3}$ since this learning rate consistently produced the best results across all models. The batch size is tuned amongst $\{16,32,64\}$ and models are trained for $10$ epochs. We report the result based on the best performance on the development set. The margin is tuned amongst $\{0.1, 0.2, 0.5\}$. All model parameters are initialized with Gaussian distributions with a mean of 0 and standard deviation of $0.1$. The L2 regularization is set to $10^{-8}$. We use a dropout of $0.5$ after the convolution or recurrent layers. A dropout of $0.8$ is set after the Coupled Attention layer in our model. Text is tokenized with NLTK's tweet tokenizer. We initialize the word embedding matrix with Glove \cite{DBLP:conf/emnlp/PenningtonSM14} trained on Twitter corpus. All words that do not appear more than $5$ times are assigned unknown tokens. All tweets are truncated at a fixed length of $10$ tokens. Early experiments found that raising the number of tokens per tweet does not improve the performance. The number of tweets per user is tuned amongst $\{10,20,50,100,150,200\}$ and reported in our experimental results.

\begin{figure}[t]
\center
 \begin{subfigure}[t]{0.22\textwidth}
  \includegraphics[width=\linewidth]{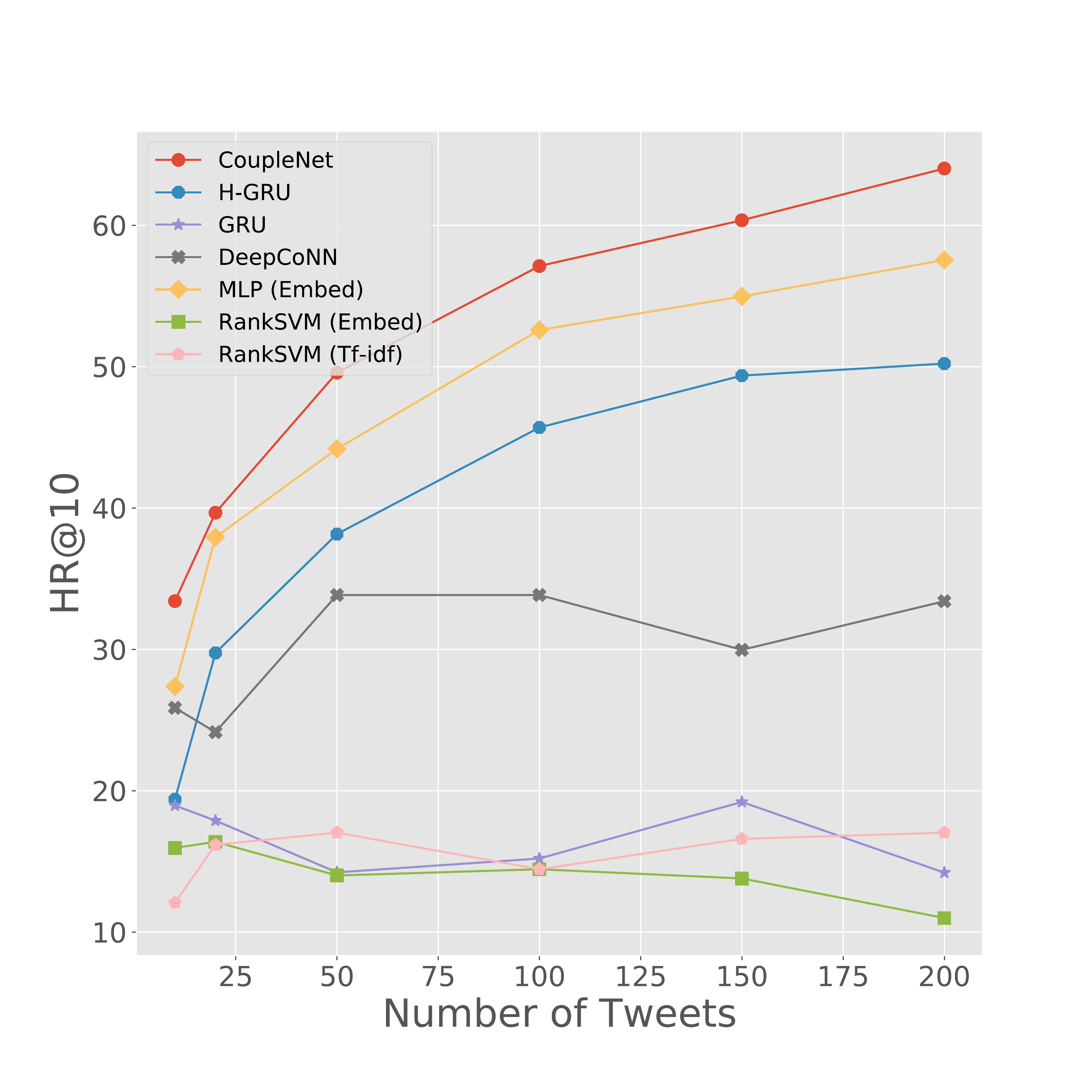}
  \caption{HR@10 Results}\label{L}
  \end{subfigure}%
\begin{subfigure}[t]{0.22\textwidth}
  \includegraphics[width=\linewidth]{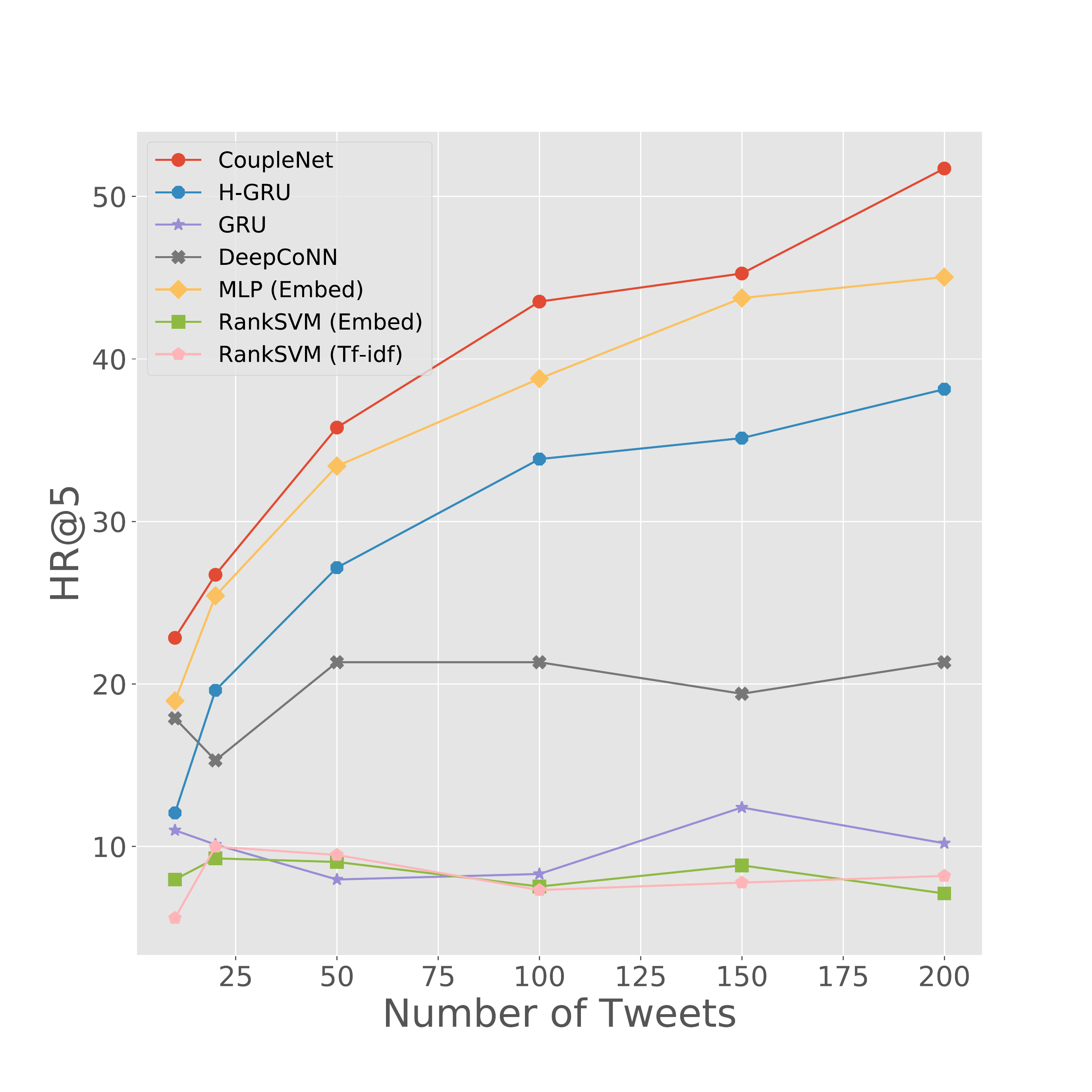}
  \caption{HR@5 Results}
\end{subfigure}%

\begin{subfigure}[t]{0.22\textwidth}
  \includegraphics[width=\linewidth]{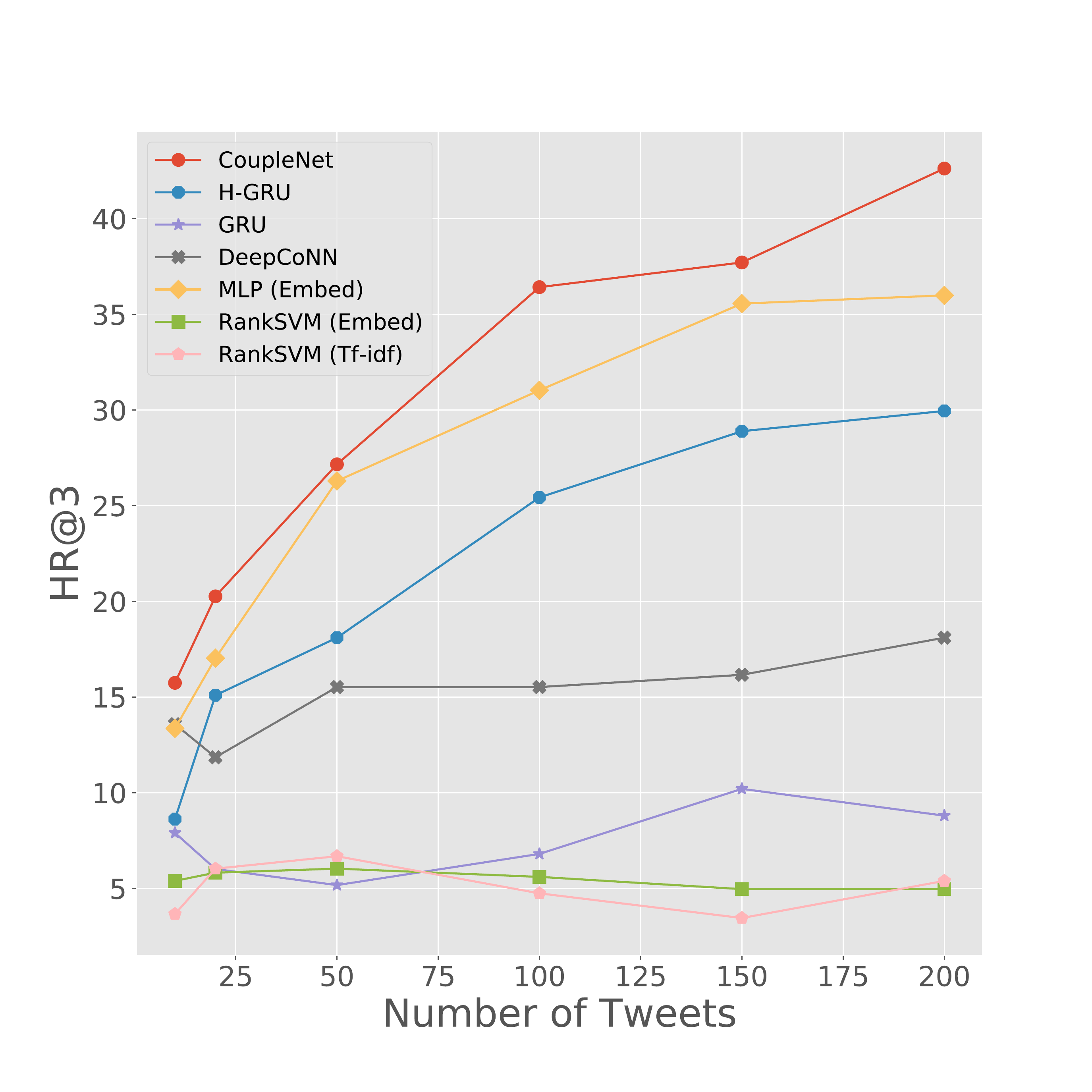}
  \caption{HR@3 Results}
\end{subfigure}
\begin{subfigure}[t]{0.22\textwidth}
  \includegraphics[width=\linewidth]{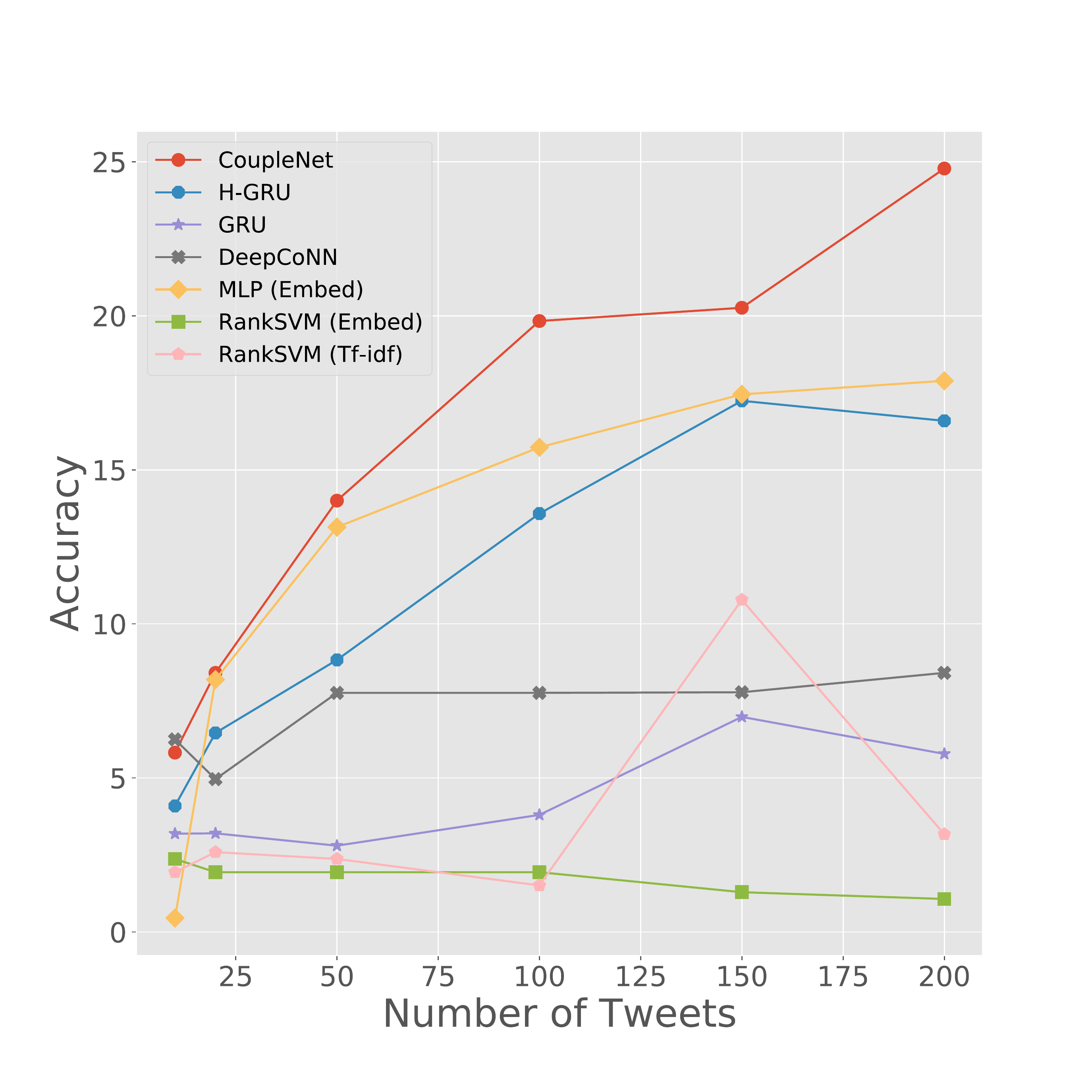}
  \caption{Accuracy Results}
\end{subfigure}%

\begin{subfigure}[t]{0.22\textwidth}
  \includegraphics[width=\linewidth]{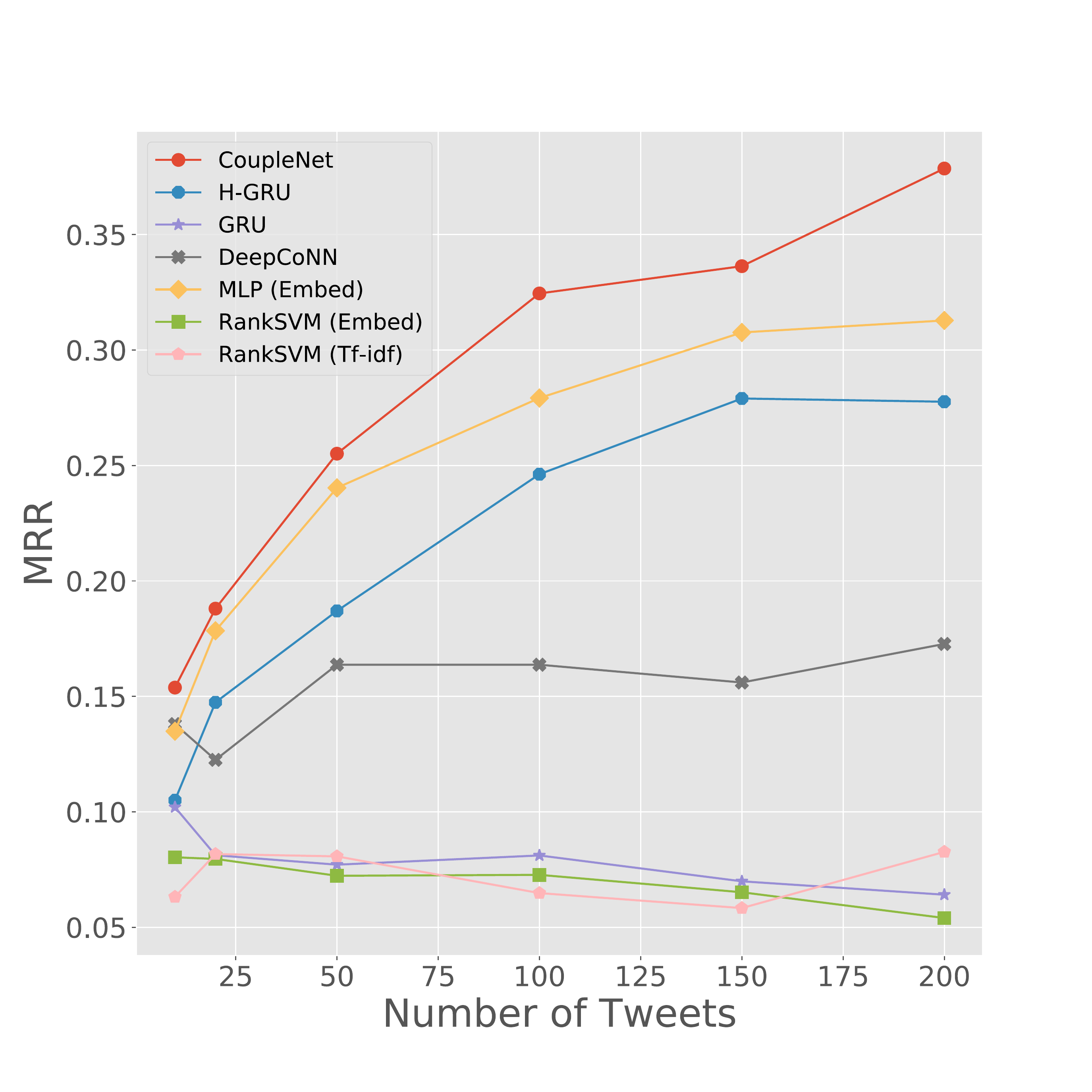}
  \caption{MRR Results}
\end{subfigure}%
\begin{subfigure}[t]{0.22\textwidth}
  \includegraphics[width=\linewidth]{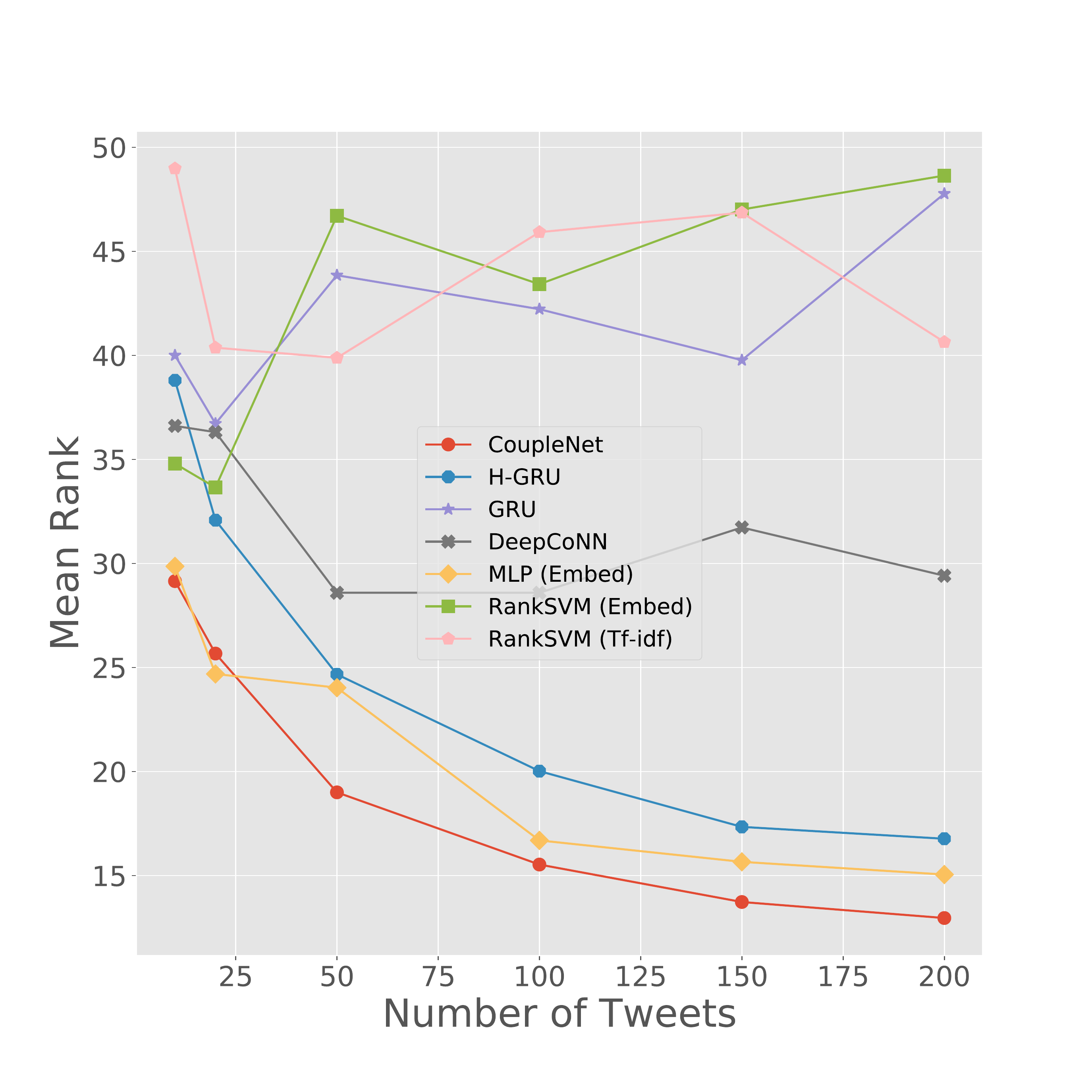}
  \caption{Mean Rank Results}
\end{subfigure}%
\caption{Experimental Results on the LoveBirds2M dataset. Results are plotted against number of tweets. \textit{Best viewed in color}. CoupleNet (\textit{red}) outperforms all baselines. }\label{exp_results}
\end{figure}

\subsection{Discussion and Analysis}
Figure \ref{exp_results} reports the experimental results on the LoveBirds2M dataset. For all baselines and evaluation metrics, we compare across different settings of $\eta$, the number of tweets per user that is used to train the model.

Firstly, we observe that \textsc{CoupleNet} significantly outperforms most of the baselines. Across most metrics, there is almost a $180\%-200\%$ relative improvement over DeepCoNN, the state-of-the-art model for item recommendation with text data. The performance improvement over the baseline GRU model is also extremely large, i.e., with a relative improvement of approximately $4$ times across all metrics. This shows that concatenating all of a user's tweets into a single document severely hurts performance. We believe that this is due to the inability of recurrent models to handle long sequences. Moreover, the DeepCoNN performs about $2$ times better than the baseline GRU model.

On the other hand, we observe that H-GRU significantly improves the baseline GRU model. In the H-GRU model, sequences are only $L=10$ long but are encoded $K$ times with shared parameters. On the other hand, the GRU model has to process $K \times L$ words, which inevitably causes performance to drop significantly. While the performance of the H-GRU model is reasonable, it is still significantly outperformed by our \textsc{CoupleNet}. We believe this is due to the incorporation of the attentional pooling layers in our model, which allows it to eliminate noise and focus on the important keywords.

A surprising and notable strong baseline is the MLP (Embed) model which outperforms DeepCoNN but still performs much worse than \textsc{CoupleNet}. On the other hand, RankSVM (Embed) performs poorly. We believe that this is attributed to the insufficiency of the linear kernel of the SVM. Since RankSVM and MLP are trained on the same features, we believe that nonlinear ReLU transformations of the MLP improve the performance significantly. Moreover, the MLP model has 2 layers, which learn different levels of abstractions. Finally, the performance of RankSVM (Tf-idf) is also poor. However, we observe that RankSVM (Tf-idf) slightly outperforms RankSVM (Embed) occasionally. While other models display a clear trend in performance with respect to the number of tweets, the performance of RankSVM (Tf-idf) and RankSVM (Embed) seem to fluctuate across the number of user tweets.

Finally, we observe a clear trend in performance gain with respect to the number of user tweets. This is intuitive because more tweets provide the model with greater insight into the user's interest and personality, allowing a better match to be made. The improvement seems to follow a logarithmic scale which suggests diminishing returns beyond a certain number of tweets. Finally, we report the time cost of \textsc{CoupleNet}. With $200$ tweets per user, the cost of training is approximately $\approx 2$ mins per epoch on a medium grade GPU. This is much faster than expected because GRUs benefit from parallism as they can process multiple tweets simultaneously.

\subsection{Ablation Study}
In this section, we study the component-wise effectiveness of \textsc{CoupleNet}. We removed layers from \textsc{CoupleNet} in order to empirically motivate the design of each component. Firstly, we switched CoupleNet to a pointwise classification model, minimizing a cross entropy loss. We found that this halves the performance. As such, we observe the importance of pairwise ranking. Secondly, we swapped cosine similarity for a MLP layer with scalar sigmoid activation (to ensure inputs lie within $[0,1]$). We also found that the performance drops significantly. Finally, we also observe that the attention layers of \textsc{CoupleNet} contribute substantially to the performance of the model. More specifically, removing both the GRU attention and coupled attention layers cause performance to drop by 13.9\%. Removing the couple attention suffers a performance degrade of $2.5\%$ while removing the GRU attention drops performance by $3.9\%$. It also seems that dropping both degrades performance more than expected (not a straightforward summation of performance degradation).
\begin{table}[htbp]
  \centering
  \small
    \begin{tabular}{lc}
    \midrule
    Model  & \multicolumn{1}{l}{HR@10} \\
    \midrule
    \textsc{CoupleNet} & 64.1 \\
    w/o couple attention & 61.6 (-2.5\%) \\
    w/o GRU attention & 60.2  (-3.9\%) \\
    w/o GRU attention and couple attention & 50.2 (-13.9\%)\\
    w/o cosine similarity & 33.8 (-30.3\%) \\
    w/o pairwise (using pointwise) & 36.1 (-28.0\%)\\
    \midrule
    \end{tabular}%
    \caption{Component-wise ablation study with $\eta=200$. }
  \label{tab:addlabel}%
\end{table}%

\subsection{Overall Quantitative Findings}
In this subsection, we describe the overall findings of our quantitative experiments.
\begin{itemize}
\item Overall, the best HR@10 score for \textsc{CoupleNet} is about $64\%$, i.e., if an application would to recommend the top $10$ prospective partners to a user, then the ground truth will appear in this list $64\%$ of the time. Moreover, the accuracy is $25\%$ (ranking out of 100 candidates) which is also reasonably high. Given the intrinsic difficulty of the problem, we believe that the performance of \textsc{CoupleNet} on this new problem is encouraging and promising. To answer \textbf{RQ1}, we believe that text-based deep learning systems for relationship recommendation are plausible. However, special care has to be taken, i.e., model selection matters.
\item The performance significantly improves when we include more tweets per user. This answers \textbf{RQ2}. This is intuitive since more tweets would enable better and more informative user representations, leading to a better matching performance.
\end{itemize}

\section{Qualitative Analysis}

In this section, we describe several insights and observations based on real\footnote{We do not explicitly report the actual user accounts in this paper because this might violate their privacy. Actual tweets are slightly modified to protect identities from search.} examples from our LoveBirds20 dataset. One key advantage of \textsc{CoupleNet} is a greater extent of explainability due to the coupled attention mechanism. More specifically, we are able to obtain which of each user's tweets contributed the most to the user representation and the overall prediction. By analyzing the attention output of user pairs, we are able to derive qualitative insights. As an overall conclusion to answer \textbf{RQ3} (which will be elaborated by in the subsequent subsections), we found that \textsc{CoupleNet} is capable of explainable recommendations if there are explicit matching signals such as user interest and demographic similarity between user pairs. Finally, we discuss some caveats and limitations of our approach.

\subsection{Mutual Interest between Couples is Captured in \textsc{CoupleNet}}
We observed the \textsc{CoupleNet} is able to capture the mutual interest between couples. Table \ref{tab:bts_table} shows an example from the LoveBirds2M dataset. In general, we found that most user pairs have noisy tweets. However, we also observed that whenever couple pairs have mutual interest, \textsc{CoupleNet} is able to assign a high attention weight to the relevant tweets. For example, in Table \ref{tab:bts_table}, both couples are fans of BTS\footnote{\url{https://en.wikipedia.org/wiki/BTS_(band)}}, a Korean pop idol group. As such, tweets related to BTS are surfaced to the top via coupled attention. In the first tweet of User 1, tweets related to two entities, \textit{seokjin} and \textit{hoseok}, are ranked high (both entities are members of the pop idol group). This ascertains that \textsc{CoupleNet} is able to, to some extent, explain why two users are matched. This also validates the usage of our coupled attention mechanism. For instance, we could infer that User1 and User2 are matched because of their mutual interest in BTS. A limitation is that it is difficult to interpret why the other tweets (such as a \textit{thank} you without much context, or \textit{supporting your family}) were ranked highly.

\begin{table}[htbp]
  \centering
  \small
    \begin{tabular}{cp{3.2cm}p{3.2cm}}
    \midrule
    \multicolumn{1}{l}{Rank} & User A & User B \\
    \midrule
    1     & i apologize to \hltwo{seokjin} and \hltwo{hoseok} 😂 & that's meant to say \hltwo{bts} but imma too tired to \\
    2     & thank you!  & more sorry for making such a mess \\
    3     & \hltwo{bts} memes mayo & i'm not sure if I shld post this \\
    4     & @user @user support your family! 😊 & the last couple of days have been shitty for me \\
    5     & welcome hun paramore! & blur pic effects are the best 😊\\
    \midrule
    \end{tabular}%
     \caption{Example of top-ranked tweets from user pair (ground truth is 1) in which mutual interests have the highest attention weight. Interest specific keywords are highlighted in red. \textsc{CoupleNet} successfully ranks this pair at the top position.}
  \label{tab:bts_table}%
\end{table}%

\subsection{\textsc{CoupleNet} Infers User Attribute and Demographic by Word Usage}
We also discovered that \textsc{CoupleNet} learns to match users with similar attributes and demographics. For example, high school students will be recommended high school students at a higher probability. Note that location, age or any other information is not provided to \textsc{CoupleNet}. In other words, user attribute and demographic are solely inferred via a user's tweets.
In Table \ref{tab:sch}, we report an example in which the top-ranked tweets (via coupled attention) are high school related tweets (homecoming, high school reception). This shows two things: (1) the coupled attention shows that the following 3 tweets were the most important tweets for prediction and (2) \textsc{CoupleNet} learns to infer user attribute and demographic without being explicitly provided with such information. We also note that both users seem to have strongly positive tweets being ranked highly in their attention scores which might hint at the role of sentiment and mood in making prediction.

\begin{table}[htbp]
  \centering
 \small
    \begin{tabular}{cp{3.2cm}p{3.2cm}}
    \midrule
    \multicolumn{1}{l}{Rank} & User C & User D \\
    \midrule
    1     & homecoming! 😁  & high school reception was a blast 😃 \\

    2     & taking meds for sports & preview will be out soon \\
    3     & so pumped for senior homecoming 😜 😜 😜& this is my life homie \\
    \midrule
    \end{tabular}%
     \caption{Example of top-ranked tweets from user pair (ground truth is 1) which are ranked by the Coupled Attention layer. \textsc{CoupleNet} places school related tweets on the top. }
  \label{tab:sch}%
\end{table}%
\subsection{\textsc{CoupleNet} Ranks Successfully Even Without Explicit Signals}
It is intuitive that not every user will post interest or demographic revealing tweets. For instance, some users might exclusively post about their emotions. When analyzing the ranking outputs of \textsc{CoupleNet}, we found that, interestingly, \textsc{CoupleNet} can successfully rank couple pairs even when there seem to be no explicit matching signal in the social profiles of both users.

Table \ref{tab:noise} shows an example where two user profiles do not share any explicit matching signals. User E and User F are a ground truth couple pair and the prediction of \textsc{CoupleNet} ranks User E with User F at the top position. The top tweets of User E and User F are mostly emotional tweets that are non-matching. Through this case, we understand that \textsc{CoupleNet} does not simply match people with similar emotions together. Notably, relationship recommendation is also a problem that humans may struggle with. Many times, the reason why two people are in a relationship may be implicit or unclear (even to humans). As such, the fact that \textsc{CoupleNet} ranks couple pairs correctly even when there is no explicit matching signals hints at its ability to go beyond simple keyword matching. In this case, we believe `hidden' (latent) patterns (such as emotions and personality) of the users are being learned and modeled in order to make recommendations. This shows that \textsc{CoupleNet} is not simply acting as a text-matching algorithm and learning features beyond that.
\begin{table}[htbp]
  \centering
\small
    \begin{tabular}{cp{3.2cm}p{3.2cm}}
    \midrule
    \multicolumn{1}{l}{Rank} & User E & User F \\
    \midrule{}
    1     & wanna be treated like a princess 😊 & can't deal with this forever 😰\\
    2     & in bed with cosy clothes and fluffy socks  & 😭 my diet is screwed \\
    3     & rt if you are currently in a mess & 😔 feel too sick \\
    4     & so much regret lmao & life is shit, home is shit \\
    5     & some girls are just so naturally pretty & still care about my grades \\
    \midrule
    \end{tabular}%
      \caption{Example of top-ranked tweets (from attention) from user pair (ground truth is 1) in which there is no explicit signal. \textsc{CoupleNet} correctly ranks this user pair at top position.}
  \label{tab:noise}%
\end{table}%

\section{Side Note, Caveats and Limitations}
While we show that our approach is capable of producing interpretable results (especially when explicit signals exist), the usefulness of its explainability may still have limitations, e.g., consider Table \ref{tab:noise} where it is clear that the results are not explainable. Firstly, there might be a complete absence of any interpretable content in two user's profiles in the first place. Secondly, explaining relationships are also challenging for humans. As such, we recommend that the outputs of \textsc{CoupleNet} to be only used as a reference. Given that a user's profile may contain easily a hundreds to thousands of tweets, one posssible use is to use this ranked list to enable more efficient analysis by humans (such as social scientist or linguists). We believe our work provides a starting point of explainable relationship recommendation.

\section{Conclusion}
We introduced a new problem of relationship recommendation. In order to construct a dataset, we employ a novel distant supervision scheme to obtain real world couples from social media. We proposed the first deep learning model for text-based relationship recommendation. Our deep learning model, \textsc{CoupleNet} is characterized by its usage of hierarchical attention-based GRUs and coupled attention layers. Performance evaluation is overall optimistic and promising. Despite huge class imbalance, our approach is able to recommend at a reasonable precision ($64\%$ at HR@10 and $25\%$ accuracy while being ranked against $100$ negative samples). Finally, our qualitative analysis shows three key findings: (1) \textsc{CoupleNet} finds mutual interests between users for match-making, (2) \textsc{CoupleNet} infers user attributes and demographics in order to make recommendations, and (3) \textsc{CoupleNet} can successfully match-make couples even when there is no explicit matching signals in their social profiles, possibly leveraging emotion and personality based latent features for prediction.

\footnotesize{
\bibliography{references}}

\begin{thebibliography}{}

\bibitem[\protect\citeauthoryear{Akehurst \bgroup et al\mbox.\egroup
  }{2011}]{DBLP:conf/ijcai/AkehurstKYPKR11}
Akehurst, J.; Koprinska, I.; Yacef, K.; Pizzato, L. A.~S.; Kay, J.; and Rej, T.
\newblock 2011.
\newblock {CCR} - {A} content-collaborative reciprocal recommender for online
  dating.
\newblock In {\em {IJCAI} 2011, Proceedings of the 22nd International Joint
  Conference on Artificial Intelligence}.

\bibitem[\protect\citeauthoryear{Arnoux \bgroup et al\mbox.\egroup
  }{2017}]{ICWSM1715681}
Arnoux, P.-H.; Xu, A.; Boyette, N.; Mahmud, J.; Akkiraju, R.; and Sinha, V.
\newblock 2017.
\newblock 25 tweets to know you: A new model to predict personality with social
  media.

\bibitem[\protect\citeauthoryear{Bahdanau, Cho, and
  Bengio}{2014}]{bahdanau2014neural}
Bahdanau, D.; Cho, K.; and Bengio, Y.
\newblock 2014.
\newblock Neural machine translation by jointly learning to align and
  translate.
\newblock {\em arXiv preprint arXiv:1409.0473}.

\bibitem[\protect\citeauthoryear{Benton, Arora, and
  Dredze}{2016}]{DBLP:conf/acl/BentonAD16}
Benton, A.; Arora, R.; and Dredze, M.
\newblock 2016.
\newblock Learning multiview embeddings of twitter users.
\newblock In {\em Proceedings of the 54th Annual Meeting of the Association for
  Computational Linguistics, {ACL} 2016}.

\bibitem[\protect\citeauthoryear{Cho \bgroup et al\mbox.\egroup
  }{2014}]{DBLP:journals/corr/ChoMGBSB14}
Cho, K.; van Merrienboer, B.; G{\"{u}}l{\c{c}}ehrse, {\c{C}}.; Bougares, F.;
  Schwenk, H.; and Bengio, Y.
\newblock 2014.
\newblock Learning phrase representations using {RNN} encoder-decoder for
  statistical machine translation.
\newblock {\em CoRR} abs/1406.1078.

\bibitem[\protect\citeauthoryear{Cobb and Kohno}{}]{DBLP:conf/www/CobbK17}
Cobb, C., and Kohno, T.
\newblock How public is my private life?: Privacy in online dating.
\newblock In {\em Proceedings of the 26th International Conference on World
  Wide Web,{WWW} 2017}.

\bibitem[\protect\citeauthoryear{Diaz, Metzler, and
  Amer{-}Yahia}{}]{DBLP:conf/sigir/DiazMA10}
Diaz, F.; Metzler, D.; and Amer{-}Yahia, S.
\newblock Relevance and ranking in online dating systems.
\newblock In {\em Proceeding of the 33rd International {ACM} {SIGIR} Conference
  on Research and Development in Information Retrieval, {SIGIR} 2010}.

\bibitem[\protect\citeauthoryear{Garimella, Weber, and
  Dal~Cin}{2014}]{garimella2014love}
Garimella, V. R.~K.; Weber, I.; and Dal~Cin, S.
\newblock 2014.
\newblock From “i love you babe” to “leave me alone”-romantic
  relationship breakups on twitter.
\newblock In {\em International Conference on Social Informatics},  199--215.
\newblock Springer.

\bibitem[\protect\citeauthoryear{Gupta \bgroup et al\mbox.\egroup
  }{2013}]{Gupta:2013:WFS:2488388.2488433}
Gupta, P.; Goel, A.; Lin, J.; Sharma, A.; Wang, D.; and Zadeh, R.
\newblock 2013.
\newblock Wtf: The who to follow service at twitter.
\newblock In {\em Proceedings of the 22Nd International Conference on World
  Wide Web}, WWW '13,  505--514.
\newblock New York, NY, USA: ACM.

\bibitem[\protect\citeauthoryear{Hancock, Toma, and
  Ellison}{2007}]{hancock2007truth}
Hancock, J.~T.; Toma, C.; and Ellison, N.
\newblock 2007.
\newblock The truth about lying in online dating profiles.
\newblock In {\em Proceedings of the SIGCHI conference on Human factors in
  computing systems},  449--452.
\newblock ACM.

\bibitem[\protect\citeauthoryear{He \bgroup et al\mbox.\egroup
  }{2017}]{He:2017:NCF:3038912.3052569}
He, X.; Liao, L.; Zhang, H.; Nie, L.; Hu, X.; and Chua, T.-S.
\newblock 2017.
\newblock Neural collaborative filtering.
\newblock In {\em Proceedings of the 26th International Conference on World
  Wide Web}, WWW '17.

\bibitem[\protect\citeauthoryear{Hu, Koren, and
  Volinsky}{2008}]{hu2008collaborative}
Hu, Y.; Koren, Y.; and Volinsky, C.
\newblock 2008.
\newblock Collaborative filtering for implicit feedback datasets.
\newblock In {\em Data Mining, 2008. ICDM'08. Eighth IEEE International
  Conference on},  263--272.
\newblock Ieee.

\bibitem[\protect\citeauthoryear{Kingma and
  Ba}{2014}]{DBLP:journals/corr/KingmaB14}
Kingma, D.~P., and Ba, J.
\newblock 2014.
\newblock Adam: {A} method for stochastic optimization.
\newblock {\em CoRR} abs/1412.6980.

\bibitem[\protect\citeauthoryear{Krzywicki \bgroup et al\mbox.\egroup
  }{2014}]{IAAI148187}
Krzywicki, A.; Wobcke, W.; Kim, Y.~S.; Cai, X.; Bain, M.; Compton, P.; and
  Mahidadia, A.
\newblock 2014.
\newblock Evaluation and deployment of a people-to-people recommender in online
  dating.

\bibitem[\protect\citeauthoryear{Luong, Pham, and
  Manning}{2015}]{luong2015effective}
Luong, M.-T.; Pham, H.; and Manning, C.~D.
\newblock 2015.
\newblock Effective approaches to attention-based neural machine translation.
\newblock {\em arXiv preprint arXiv:1508.04025}.

\bibitem[\protect\citeauthoryear{Maldeniya \bgroup et al\mbox.\egroup
  }{2017}]{ICWSM1715634}
Maldeniya, D.; Varghese, A.; Stuart, T.; and Romero, D.
\newblock 2017.
\newblock The role of optimal distinctiveness and homophily in online dating.

\bibitem[\protect\citeauthoryear{Masden and
  Edwards}{2015}]{Masden:2015:URC:2702123.2702417}
Masden, C., and Edwards, W.~K.
\newblock 2015.
\newblock Understanding the role of community in online dating.
\newblock In {\em Proceedings of the 33rd Annual ACM Conference on Human
  Factors in Computing Systems}, CHI '15,  535--544.
\newblock New York, NY, USA: ACM.

\bibitem[\protect\citeauthoryear{Nagarajan and
  Hearst}{2009}]{DBLP:conf/icwsm/NagarajanH09}
Nagarajan, M., and Hearst, M.~A.
\newblock 2009.
\newblock An examination of language use in online dating profiles.
\newblock In {\em Proceedings of the Third International Conference on Weblogs
  and Social Media, {ICWSM} 2009, San Jose, California, USA, May 17-20, 2009}.

\bibitem[\protect\citeauthoryear{Pennington, Socher, and
  Manning}{2014}]{DBLP:conf/emnlp/PenningtonSM14}
Pennington, J.; Socher, R.; and Manning, C.~D.
\newblock 2014.
\newblock Glove: Global vectors for word representation.
\newblock In {\em Proceedings of the 2014 Conference on Empirical Methods in
  Natural Language Processing, {EMNLP}}.

\bibitem[\protect\citeauthoryear{Severyn and
  Moschitti}{2015}]{DBLP:conf/sigir/SeverynM15}
Severyn, A., and Moschitti, A.
\newblock 2015.
\newblock Learning to rank short text pairs with convolutional deep neural
  networks.
\newblock In {\em Proceedings of the 38th International {ACM} {SIGIR}
  Conference on Research and Development in Information Retrieval}.

\bibitem[\protect\citeauthoryear{Tay, Anh~Tuan, and
  Hui}{2018}]{Tay:2018:LRM:3178876.3186154}
Tay, Y.; Anh~Tuan, L.; and Hui, S.~C.
\newblock 2018.
\newblock Latent relational metric learning via memory-based attention for
  collaborative ranking.
\newblock In {\em Proceedings of the 2018 World Wide Web Conference}, WWW '18,
  729--739.
\newblock Republic and Canton of Geneva, Switzerland: International World Wide
  Web Conferences Steering Committee.

\bibitem[\protect\citeauthoryear{Tay \bgroup et al\mbox.\egroup
  }{2017}]{DBLP:conf/sigir/TayPLH17}
Tay, Y.; Phan, M.~C.; Luu, A.~T.; and Hui, S.~C.
\newblock 2017.
\newblock Learning to rank question answer pairs with holographic dual {LSTM}
  architecture.
\newblock In {\em Proceedings of the 40th International {ACM} {SIGIR}
  Conference on Research and Development in Information Retrieval, 2017}.

\bibitem[\protect\citeauthoryear{Tay, Tuan, and Hui}{2017}]{tay2017cross}
Tay, Y.; Tuan, L.~A.; and Hui, S.~C.
\newblock 2017.
\newblock Cross temporal recurrent networks for ranking question answer pairs.
\newblock {\em arXiv preprint arXiv:1711.07656}.

\bibitem[\protect\citeauthoryear{Tay, Tuan, and
  Hui}{2018}]{DBLP:journals/corr/abs-1801-09251}
Tay, Y.; Tuan, L.~A.; and Hui, S.~C.
\newblock 2018.
\newblock Multi-pointer co-attention networks for recommendation.
\newblock {\em CoRR} abs/1801.09251.

\bibitem[\protect\citeauthoryear{Toma and
  Hancock}{2010}]{Toma:2010:RLL:1718918.1718921}
Toma, C.~L., and Hancock, J.~T.
\newblock 2010.
\newblock Reading between the lines: Linguistic cues to deception in online
  dating profiles.
\newblock In {\em Proceedings of the CSCW, 2010}.

\bibitem[\protect\citeauthoryear{Tu \bgroup et al\mbox.\egroup
  }{2014}]{Tu:2014:ODR:2567948.2579240}
Tu, K.; Ribeiro, B.; Jensen, D.; Towsley, D.; Liu, B.; Jiang, H.; and Wang, X.
\newblock 2014.
\newblock Online dating recommendations: Matching markets and learning
  preferences.
\newblock In {\em Proceedings of the 23rd International Conference on World
  Wide Web}, WWW '14 Companion,  787--792.
\newblock New York, NY, USA: ACM.

\bibitem[\protect\citeauthoryear{Wei \bgroup et al\mbox.\egroup
  }{2017}]{Wei:2017:BWP:3018661.3018717}
Wei, H.; Zhang, F.; Yuan, N.~J.; Cao, C.; Fu, H.; Xie, X.; Rui, Y.; and Ma,
  W.-Y.
\newblock 2017.
\newblock Beyond the words: Predicting user personality from heterogeneous
  information.
\newblock In {\em Proceedings of the Tenth ACM International Conference on Web
  Search and Data Mining}, WSDM '17,  305--314.
\newblock New York, NY, USA: ACM.

\bibitem[\protect\citeauthoryear{Xia \bgroup et al\mbox.\egroup
  }{2014}]{ICWSM148061}
Xia, P.; Jiang, H.; Wang, X.; Chen, C.; and Liu, B.
\newblock 2014.
\newblock Predicting user replying behavior on a large online dating site.

\bibitem[\protect\citeauthoryear{Xia \bgroup et al\mbox.\egroup
  }{2015}]{Xia:2015:RRS:2808797.2809282}
Xia, P.; Liu, B.; Sun, Y.; and Chen, C.
\newblock 2015.
\newblock Reciprocal recommendation system for online dating.
\newblock In {\em Proceedings of the 2015 IEEE/ACM International Conference on
  Advances in Social Networks Analysis and Mining 2015}.

\bibitem[\protect\citeauthoryear{Yang \bgroup et al\mbox.\egroup
  }{2016}]{yang2016hierarchical}
Yang, Z.; Yang, D.; Dyer, C.; He, X.; Smola, A.~J.; and Hovy, E.~H.
\newblock 2016.
\newblock Hierarchical attention networks for document classification.

\bibitem[\protect\citeauthoryear{Zhang \bgroup et al\mbox.\egroup
  }{2018}]{zhang2018neurec}
Zhang, S.; Yao, L.; Sun, A.; Wang, S.; Long, G.; and Dong, M.
\newblock 2018.
\newblock Neurec: On nonlinear transformation for personalized ranking.
\newblock {\em arXiv preprint arXiv:1805.03002}.

\bibitem[\protect\citeauthoryear{Zhang, Yao, and Sun}{2017}]{zhang2017deep}
Zhang, S.; Yao, L.; and Sun, A.
\newblock 2017.
\newblock Deep learning based recommender system: A survey and new
  perspectives.
\newblock {\em arXiv preprint arXiv:1707.07435}.

\bibitem[\protect\citeauthoryear{Zheng, Noroozi, and Yu}{2017}]{zheng2017joint}
Zheng, L.; Noroozi, V.; and Yu, P.~S.
\newblock 2017.
\newblock Joint deep modeling of users and items using reviews for
  recommendation.
\newblock In {\em Proceedings of the Tenth ACM International Conference on Web
  Search and Data Mining},  425--434.
\newblock ACM.

\end{thebibliography}
\bibliographystyle{aaai}
\end{document}